# Evaluation of Protein Structural Models Using Random Forests

Renzhi Cao[1], Taeho Jo[1,2], Jianlin Cheng[1*]

[1]Department of Computer Science, University of Missouri, Columbia, MO 65211, USA

[2]Department of Biological Chemistry, University of Michigan, Ann Arbor, MI, 48109, USA

*Corresponding author: chengji@missouri.edu

## Abstract

**Protein structure prediction has been a "grand challenge" problem in the structure biology over the last few decades. Protein quality assessment plays a very important role in protein structure prediction. In the paper, we propose a new protein quality assessment method which can predict both local and global quality of the protein 3D structural models. Our method uses both multi and single model quality assessment method for global quality assessment, and uses chemical, physical, geo-metrical features, and global quality score for local quality assessment. CASP9 targets are used to generate the features for local quality assessment. We evaluate the performance of our local quality assessment method on CASP10, which is comparable with two stage-of-art QA methods based on the average absolute distance between the real and predicted distance. In addition, we blindly tested our method on CASP11, and the good performance shows that combining single and multiple model quality assessment method could be a good way to improve the accuracy of model quality assessment, and the random forest technique could be used to train a good local quality assessment model.**

# Introduction

The protein structure prediction has been defined as one of the grand challenges problem in bioinformatics and computational biology, and still not solved over the last several decades [1]. With the development of computer, a huge quantity of computational methods has been generated to predict the protein tertiary structure from the amino acid [2-8]. These methods are mainly divided into the following classes [3,9]: the template-based methods [3,4,7], which uses the known protein structure determined by biology experiment as template to predict the structure of new query protein sequence; the template free methods [7,10,11], which try to predict the protein structure from the amino acid sequence without directly using any known protein structures; hybrid methods [2,3,9,12,13], which takes advantage of the previous two different methods, and generate more accurate protein structures. For all of these different protein structure prediction methods, there is one common important and unsolved problem: which predicted protein structure model is closer to the truth without knowing the native protein structure? That is the protein quality assessment (QA) problem. The model mentioned in this paper represents protein 3D structural model. The protein quality assessment is very useful for selecting the good models from the model pool, to refine the predicted models, and etc [3]. In general, there are two different qualities for the predicted protein model: local and global quality. The global quality score shows how close of the predicted model to the native structure, and the local quality score shows how close of each residues in the predicted model to the native structure. There are two different strategies to evaluate the quality of a predicted model [2]: multi-model methods [3,4,14-18] and single-model methods [5,19-23]. Multi-model methods use pairwise or clustering technique to compare the similarity of each model against all others, and then define the good model as the one which is most similar to all other model. This method works well when a large proportion of the model pool has good quality, such as the case of easy template-based modeling. However, it may fail when a significant portion of low quality modes are dominating the pairwise model comparison since they are very similar to each other [2]. The single-model methods [19-23] predict the predicted protein model's quality without using the information of other models.

In this paper, we introduce a hybrid method to predict the global quality score of the input models, and one random forest based method to predict the local quality score. For the global quality assessment, the pairwise score

[14] is generated from the model pool, and an improved version of model evaluator [24] (model check2 score) is also calculated. Either the pair score, or model check2 score is used as the final global quality score. For the local quality assessment, the local features are generated from physical, chemical, and geometrical respective[25] using sliding window size 15 centered on a target residue, and also using the global features from the whole model. Random forest is an ensemble classification which uses tree-structured classifiers. Random forest grows a large number of decision trees, trains them applying the general technique of bootstrap aggregating (bagging). The predictions are determined by majority vote of trees. Because the ensemble reduces variance, random forest is robust to change in data, irrelevant features, and unbalanced class distribution. Random forest showed excellent performance in broad classification tasks[26], which is generally comparable to that of other ensemble classifiers such as AdaBoost [13] or traditional machine learning classification algorithm such as SVM [2,5] or Deep Learning Networks [27].

The rest of the paper is organized as follows: in the methods section, we will describe the method we use to predict the local and global quality of the input models; in the discussion section, we will evaluate the performance of our method on CASP10, and compare our method with other methods, and then discuss our method's performance; in the conclusion section, we summarize our work.

# Results and discussion

The global quality assessment is tested on CASP10 targets, and also blindly benchmarked on CASP11 targets. We trained our local quality assessment model based on CASP9 targets, and five cross validation is used for training the random forest model.

We first evaluate the performance of our global quality assessment method, and then evaluate the performance of our local quality assessment.

**Evaluation of global quality predictions**

Our global quality assessment method is a hybrid method, and we choose the maximum pairwise GDT-TS score 0.2 as the threshold to decide which method should be used for the input target, either pairwise method or model check2. Figure 1 shows the average correlation of pairwise method for CASP10 stage1 and stage2 targets with different maximum pairwise score. The x-axis describes maximum pairwise score threshold for all targets. The y-axis shows the average correlation of all targets within the x threshold. This figures shows that as the maximum pairwise score decreases, the performance of pairwise method also decreases for the CASP10 stage1 targets, and the performance decreases a lot between the maximum pairwise score threshold 0.25 to 0.3 on CASP10 stage2 targets. Considering the performance of pairwise method on CASP10 targets, finally we decide to use the threshold 0.2 to decide whether we use pairwise method.

**Table 1** and **Table 2** show the average correlation and loss of our global quality assessment method on CASP10 stage1 and stage2 targets respectively. We also include the performance of other top groups' method, such as ModFOLDclust2 method which is based on clustering technique, and ProQ2 method which is one of the best single quality assessment method on CASP10. As the table shows, the pairwise method ModFOLDclust2 performs better than the single quality assessment ProQ2 method from the respective of average correlation, overall correlation and loss. However, the difference between them becomes less on stage2, and the loss between ModFOLDclust2 and ProQ2 is the same on stage2. This tell us that the single QA method has the similar ability to find out the best model out of the model pool comparing with the pairwise method. As our MULTICOM-REFINE server, we take the advantage of both single and pairwise QA method, and the results show that our method gets better performance on both stage1 and stage2 comparing with the state-of-art QA methods, e.g, the average correlation of our method on stage1 is better than ProQ2, and the loss on stage1 is less than ModFOLDclust2, and our method has the biggest average correlation comparing with other two methods on stage2.

As we know that pairwise model assessment methods worked better when a large portion of models in the pool were of good quality, whereas single-model quality assessment methods performed better on some hard targets when only a small portion of models in the pool were of reasonable quality [2], so it would be interest to see the performance of different methods on the human targets of CASP10. **Table 3** and **Table 4** show the performance of our global quality assessment method on stage1 and stage2 of human targets respectively, and we also include the other group's method like ModFOLDclust2 and ProQ2 method. Indeed, we can see from the table, the pairwise and single QA method gets similar performance on the human targets. Moreover, as we can see from **Table 3**, the average correlation of ProQ2 on human targets is 0.58, which is the same as the pairwise method ModFOLDclust2 (the average correlation on stage1 of all targets is 0.68). The average correlation of MULTICOM-REFINE on stage1 is similar to other methods, and the loss is the smallest between ProQ2 and ModFOLDclust2 method. Similar pattern can be found on stage2 for MULTICOM-REFINE. Overall, MULTICOM-REFINE gets better performance on both stage1 and stage2 of CASP10. In addition, our method attends CASP11, and we also show the performance of our method on stage1 and stage2 for all CASP11 targets comparing with ModFOLDclust2 and ProQ2 method at **Table 5** and **Table 6**. We can find out from **Table 5** that our method is better than state-of-art single model QA method ProQ2 based on average correlation or loss, and is better than ModFOLDclust2 based on the average correlation also. **Table 6** shows that our method gets similar performance comparing with ModFOLDclust2 on stage2, and better that ProQ2 method.

**Evaluation of local quality predictions**

We evaluate the performance of our local quality assessment method on CASP10 targets on stage1 and stage2. In order to make comparison with other state-of-art local quality assessment methods, we also evaluate the ProQ2 (single model quality assessment tool) and ModFOLDclust2 (multi-model quality assessment tool). In order to evaluate the performance, we calculate the absolute distance difference between real and predicted local distance for each residue. Smaller difference means higher accurate of the predictions. **Figure 2** and **Figure 3** shows the relationship of the real and predicted distance of 98 CASP10 targets on stage1 and stage2 respectively. The x-axis is the real distance between the native and model, which is divided in to 20 bins. The y-axis shows the average of absolute difference between real and predicted distance in each real distance bin. From these two figures, we can see that our MULTICOM-REFINE's new local quality assessment method based on random forest is comparable with the state-of-art local quality assessment method. Especially when the real distance is less than 7 angstrom, the average absolute difference between real and our prediction is very close to the other two methods. Our method even has smaller average absolute difference comparing with the other two methods, e.g, for 98 CASP10 targets on stage2, the average absolute difference when the real distance is 3 angstrom is 0.61, 0.88, and 1.33 for MULTICOM-REFINE, ProQ2, and ModFOLDclust2 respectively. We may also notice that when the real distance is larger than 7, the clustering method ModFOLDclust2 works better than single model quality assessment ProQ2 and our method. There could be two reasons that our method doesn't perform well when the real distance is large. The first reason is that our method tends to predict smaller distance, and we set a threshold 15 for all predictions, so that there is no prediction larger than 15 for our method. The second reason could be in the training data, we involve more accurate models, so that our trained model is more likely to predict small distance for each residue.

## Conclusions

In this paper, we describe a local and global quality assessment method. Our global quality assessment method takes advantage of pairwise and single-model QA method, and generates better performance comparing with pairwise method and the single-model QA method. We also evaluate the performance of our method on hard target, it shows that our method consistently gets better performance comparing with two state-of-art quality assessment method ProQ2 and ModFOLDclust2. For the local quality assessment part, we evaluate our method's performance on CASP10 targets, and it shows that our method is comparable with the other two state-of-art model quality assessment method. In the future, we plan to add more training data to improve the accuracy of our local quality assessment method, and consider influence of domains for training the local quality assessment model from the protein structure. Also for the global quality assessment method, we plan to rigorously test it on larger dataset, and find other better way to combine the pairwise and single-model methods. Overall, we believe our method performs well on the CASP10 targets and also has good performance on CASP11, which shows the potential of combining pairwise and single-model method, and also there are a lot of improvements for the local quality assessment method. The web server is built for public use of our method at: http://calla.rnet.missouri.edu/rfqa/.

# Methods

The global quality assessment of this paper is a hybrid method, and the local quality assessment of this paper is a single model method, which is trained by random forest technique on CASP9 targets. The method to predict the global and local quality scores are introduced in the following sections.

**Global quality assessment method**

First of all, the improved version of model evaluator model check2 is used to calculate the score for each input model. Second, while the number of models is larger than one, the pairwise method is applied to the input model pool [14]. The GDT-TS score of each model against all other model is calculated using TM-score [28], and the average GDT-TS score is calculated for each model as the quality of that model. Finally, the maximum of GDT-TS score among the model pool is used to decide which score to be used as the global quality score as the model pool. The pairwise score is used when the maximum GDT-TS score is larger than 0.2, otherwise, model check2 score is used.

**Local quality features preparation**

All CASP9 targets are used to generate the local quality features. There are two different types of local quality features, one is global features coming from the quality of the model, and the other is local features coming from the amino acids with sliding window size 15. For each residue, we generate a feature set for making local quality assessment. In total, 4,719,526 feature sets are generated from CASP9 targets. According to the real quality of each feature set, we divide the data into 5 classes. For example, the first class is that all feature sets with the real quality from 0 to 0.2. We randomly select 10,000 feature sets from each class due to the time complexity of training random forest model with large training data set. The RandomForest package in R [11] is used for training the random forest model. The global features includes the difference between secondary structure and solvent accessibility predicted by Spine X [29] and SSpro4 [30] from the protein sequence and that of a model parsed by DSSP [31], the pairwise Euclidean distance score which is calculated by the average Euclidean distance of the model for all pairwise amino acid pairs divided by the same distance of the extended structure for the model, secondary structure penalty score which is calculated from the mismatch of helix and sheet between the predicted secondary structure and the one parsed from the model [25], surface polar score which is calculated by the fractional area of exposed nonpolar residues [25], weighted exposed area score which is the weighted exposed area divide by the whole area [25], total surface area score which is the total surface area divided by the whole area [25]. The local features for each amino acid is coming from the fragment with sliding window size 15, including the amino acids encoded by a 20-digit vector of 0 and 1, secondary structure difference, pairwise Euclidean distance score, secondary structure penalty score, surface polar score, weighted exposed area score, and total area score generated from the fragment.

**Train a model for local quality assessment by random forest**

We divided the 10,000 feature sets which were explained above into 10 equal-size subsets for 10-fold cross validation. Nine subsets were used for training and the remaining subset was used for validation. A number of feature sets were randomly selected from each subset for constructing decision trees and standard decision tree training algorithm was applied. After training, the average probability predicted by these trees was calculated as the local quality score. This procedure was repeated 10 times and the sensitivity and specificity were computed across the 10 trials.

# References

1   Zhang, Y. & Skolnick, J. SPICKER: a clustering approach to identify near-native protein folds. *J Comput Chem* **25**, 865 - 871 (2004).
2   Cao, R., Wang, Z. & Cheng, J. Designing and evaluating the MULTICOM protein local and global model quality prediction methods in the CASP10 experiment. *BMC structural biology* **14**, 13 (2014).
3   Cao, R., Bhattacharya, D., Adhikari, B., Li, J. & Cheng, J. Large-scale model quality assessment for improving protein tertiary structure prediction. *Bioinformatics* **31**, i116-i123 (2015).
4   Cao, R., Bhattacharya, D., Adhikari, B., Li, J. & Cheng, J. Massive integration of diverse protein quality assessment methods to improve template based modeling in CASP11. *Proteins: Structure, Function, and Bioinformatics* (2015).
5   Cao, R., Wang, Z., Wang, Y. & Cheng, J. SMOQ: a tool for predicting the absolute residue-specific quality of a single protein model with support vector machines. *BMC bioinformatics* **15**, 120 (2014).
6   Li, J., Cao, R. & Cheng, J. in *Proceedings of the 6th ACM Conference on Bioinformatics, Computational Biology and Health Informatics.*  537-537 (ACM).
7   Adhikari, B., Bhattacharya, D., Cao, R. & Cheng, J. CONFOLD: Residue-residue contact-guided ab initio protein folding. *Proteins: Structure, Function, and Bioinformatics* (2015).
8   Li, J. *et al.* in *Protein Structure Prediction*     29-41 (Springer, 2014).
9   Wang, Z., Cao, R. & Cheng, J. Three-level prediction of protein function by combining profile-sequence search, profile-profile search, and domain co-occurrence networks. *BMC bioinformatics* **14**, S3 (2013).
10  Cao, R. & Cheng, J. Integrated protein function prediction by mining function associations, sequences, and protein-protein and gene-gene interaction networks. *Methods* (2015).

11. Liaw, A. & Wiener, M. Classification and regression by randomForest. *R news* **2**, 18-22 (2002).
12. Jones, D. T., Taylort, W. & Thornton, J. M. A new approach to protein fold recognition. (1992).
13. Yang, Y. & Zhou, Y. Ab initio folding of terminal segments with secondary structures reveals the fine difference between two closely related all-atom statistical energy functions. *Protein Science* **17**, 1212-1219 (2008).
14. Khan, I. K., Wei, Q., Chitale, M. & Kihara, D. PFP/ESG: automated protein function prediction servers enhanced with Gene Ontology visualization tool. *Bioinformatics*, btu646 (2014).
15. McGuffin, L. J. The ModFOLD server for the quality assessment of protein structural models. *Bioinformatics* **24**, 586-587 (2008).
16. McGuffin, L. J. Prediction of global and local model quality in CASP8 using the ModFOLD server. *Proteins: Structure, Function, and Bioinformatics* **77**, 185-190 (2009).
17. Wang, Q., Vantasin, K., Xu, D. & Shang, Y. MUFOLD-WQA: A new selective consensus method for quality assessment in protein structure prediction. *Proteins: Structure, Function, and Bioinformatics* (2011).
18. McGuffin, L. J. & Roche, D. B. Rapid model quality assessment for protein structure predictions using the comparison of multiple models without structural alignments. *Bioinformatics* **26**, 182-188 (2010).
19. Wallner, B. & Elofsson, A. Can correct protein models be identified? *Protein Science* **12**, 1073-1086 (2003).
20. Ray, A., Lindahl, E. & Wallner, B. Improved model quality assessment using ProQ2. *BMC bioinformatics* **13**, 224 (2012).
21. McGuffin, L. J., Bryson, K. & Jones, D. T. The PSIPRED protein structure prediction server. *Bioinformatics* **16**, 404-405 (2000).
22. Benkert, P., Tosatto, S. C. & Schomburg, D. QMEAN: A comprehensive scoring function for model quality assessment. *Proteins: Structure, Function, and Bioinformatics* **71**, 261-277 (2008).
23. Fontana, P., Cestaro, A., Velasco, R., Formentin, E. & Toppo, S. Rapid annotation of anonymous sequences from genome projects using semantic similarities and a weighting scheme in gene ontology. *PLoS One* **4**, e4619 (2009).
24. Wang, Z., Tegge, A. N. & Cheng, J. Evaluating the absolute quality of a single protein model using structural features and support vector machines. *Proteins* **75**, 638-647, doi:10.1002/prot.22275 (2009).
25. Kabsch, W. & Sander, C. Dictionary of protein secondary structure: pattern recognition of hydrogen-bonded and geometrical features. *Biopolymers* **22**, 2577 - 2637 (1983).
26. Jo, T. & Cheng, J. Improving protein fold recognition by random forest. *BMC bioinformatics* **15**, S14 (2014).
27. Jo, T., Hou, J., Eickholt, J. & Cheng, J. Improving protein fold recognition by deep learning networks. *Sic. Rep* **5**, 17573 (2015).
28. Zhang, Y. & Skolnick, J. Scoring function for automated assessment of protein structure template quality. *Proteins: Structure, Function, and Bioinformatics* **57**, 702-710 (2004).
29. Faraggi, E., Zhang, T., Yang, Y., Kurgan, L. & Zhou, Y. SPINE X: improving protein secondary structure prediction by multistep learning coupled with prediction of solvent accessible surface area and backbone torsion angles. *Journal of computational chemistry* **33**, 259-267 (2012).
30. Cheng, J., Randall, A. Z., Sweredoski, M. J. & Baldi, P. SCRATCH: a protein structure and structural feature prediction server. *Nucleic Acids Research* **33**, W72-W76 (2005).
31. Kabsch, W. & Sander, C. Dictionary of protein secondary structure: pattern recognition of hydrogen-bonded and geometrical features. *Biopolymers* **22**, 2577-2637 (1983).

# Acknowledgements

The work was partially supported by an NIH grant (R01GM093123) to JC.

# Author Contributions

JC conceived and designed the method and the system. RC, TJ implemented the method, built the system, carried out the CASP experiments. RC, TJ, JC evaluated and analyzed data. RC, TJ wrote the manuscript. All the authors approved the manuscript.

# Competing financial interests

The authors declared no competing financial interests.

# Figure legends

## Figure 1

The average correlation of pairwise method for CASP10 stage1 and stage2 targets with different maximum pairwise score

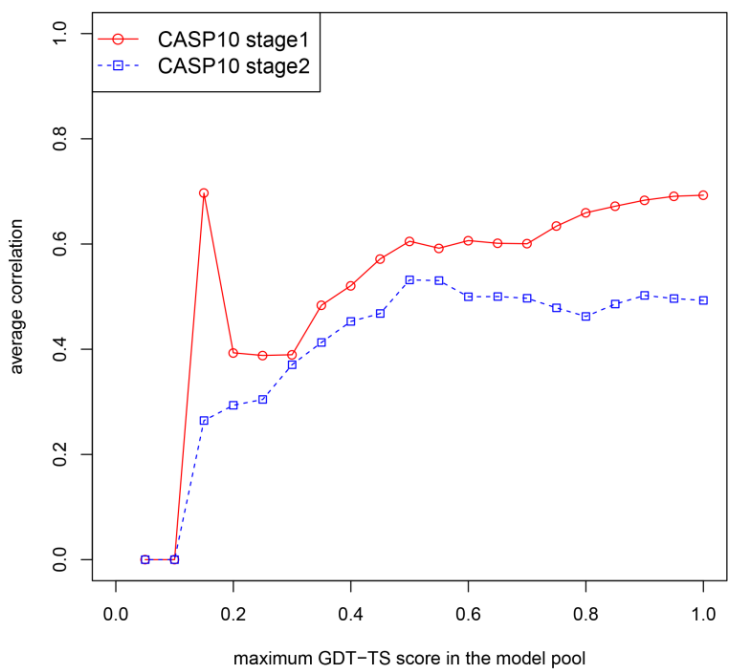

# Figure 2

The absolute difference between real and predicted distance against the real distance in 20 bins for 98 CASP10 targets on stage1.

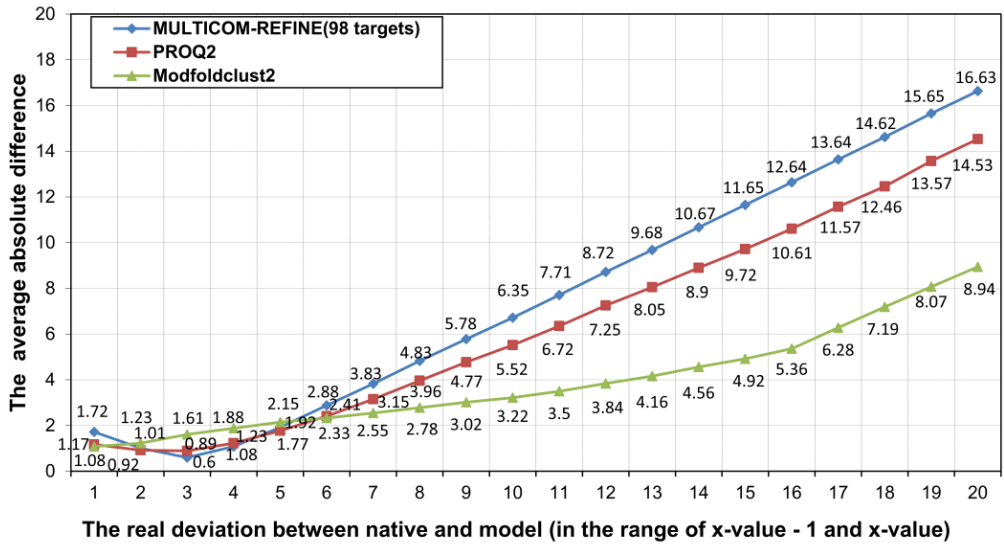

# Figure 3

The absolute difference between real and predicted distance against the real distance in 20 bins for 98 CASP10 targets on stage2.

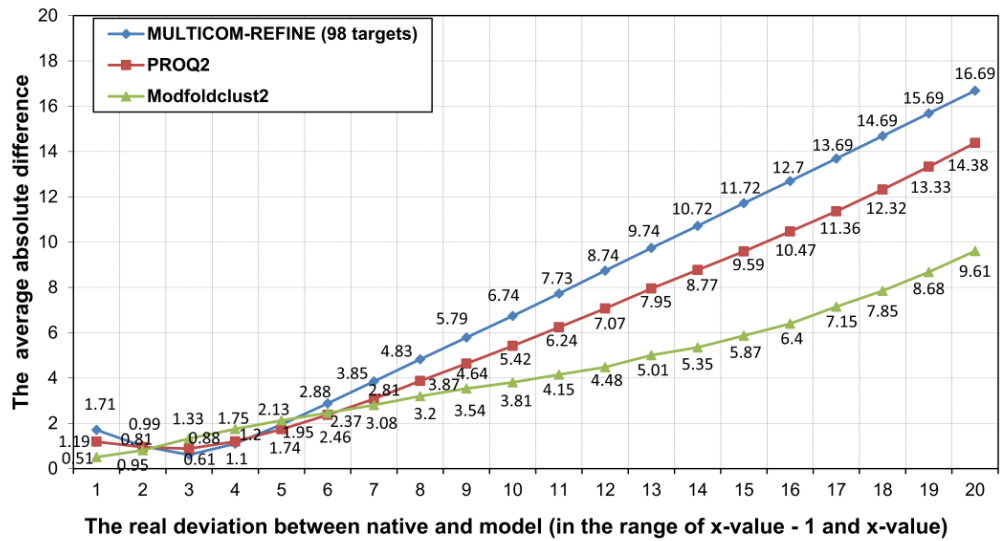

**Table 1.** The average correlation (Ave. Corr.), overall correlation (Over. Corr.), and average GDT-TS loss (Ave. loss) of MULTICOM-REFINE server, proq2, and ModFOLDclust2 on Stage1 of CASP10.

| Stage1 of CASP10 | Ave. Corr. | Over. Corr. | Ave. loss |
| --- | --- | --- | --- |
| MULTICOM-REFINE | 0.68 | 0.81 | 0.05 |
| Proq2 | 0.58 | 0.61 | 0.07 |
| ModFOLDclust2 | 0.68 | 0.83 | 0.06 |

**Table 2.** The average correlation (Ave. Corr.), overall correlation (Over. Corr.), and average GDT-TS loss (Ave. loss) of MULTICOM-REFINE server, ProQ2, and ModFOLDclust2 on Stage2 of CASP10.

| Stage2 of CASP10 | Ave. Corr. | Over. Corr. | Ave. loss |
| --- | --- | --- | --- |
| MULTICOM-REFINE | 0.48 | 0.83 | 0.05 |
| ProQ2 | 0.42 | 0.60 | 0.05 |
| ModFOLDclust2 | 0.45 | 0.83 | 0.05 |

**Table 3.** The average correlation (Ave. Corr.), overall correlation (Over. Corr.), and average GDT-TS loss (Ave. loss) of MULTICOM-REFINE server, ProQ2, and ModFOLDclust2 on Stage1 of all human targets of CASP10.

| Stage1 of CASP10 | Ave. Corr. | Over. Corr. | Ave. loss |
| --- | --- | --- | --- |
| MULTICOM-REFINE | 0.59 | 0.81 | 0.06 |
| ProQ2 | 0.58 | 0.52 | 0.08 |
| ModFOLDclust2 | 0.58 | 0.86 | 0.08 |

**Table 4.** The average correlation (Ave. Corr.), overall correlation (Over. Corr.), and average GDT-TS loss (Ave. loss) of MULTICOM-REFINE server, ProQ2, and ModFOLDclust2 on Stage2 of all human targets of CASP10.

| Stage2 of CASP10 | Ave. Corr. | Over. Corr. | Ave. loss |
| --- | --- | --- | --- |
| MULTICOM-REFINE | 0.50 | 0.85 | 0.06 |
| ProQ2 | 0.41 | 0.48 | 0.06 |
| ModFOLDclust2 | 0.46 | 0.87 | 0.05 |

**Table 5.** The average correlation (Ave. Corr.), overall correlation (Over. Corr.), and average GDT-TS loss (Ave. loss) of MULTICOM-REFINE server, ProQ2, and ModFOLDclust2 on Stage1 of all human targets of CASP11.

| Stage1 of CASP11 | Ave. Corr. | Over. Corr. | Ave. loss |
|---|---|---|---|
| MULTICOM-REFINE | 0.80 | 0.93 | 0.05 |
| ProQ2 | 0.64 | 0.79 | 0.09 |
| ModFOLDclust2 | 0.74 | 0.95 | 0.05 |

**Table 6.** The average correlation (Ave. Corr.), overall correlation (Over. Corr.), and average GDT-TS loss (Ave. loss) of MULTICOM-REFINE server, ProQ2, and ModFOLDclust2 on Stage2 of all human targets of CASP11.

| Stage2 of CASP11 | Ave. Corr. | Over. Corr. | Ave. loss |
|---|---|---|---|
| MULTICOM-REFINE | 0.57 | 0.95 | 0.07 |
| ProQ2 | 0.37 | 0.76 | 0.06 |
| ModFOLDclust2 | 0.56 | 0.95 | 0.07 |